\documentclass[10pt,twocolumn,letterpaper]{article}

\usepackage[final]{cvpr}      
\usepackage{subcaption}
\usepackage{url}

\usepackage[dvipsnames]{xcolor}

\definecolor{cvprblue}{rgb}{0.21,0.49,0.74}
\usepackage[pagebackref,breaklinks,colorlinks,citecolor=cvprblue]{hyperref}

\def\paperID{*****} 
\def\confName{CVPR}
\def\confYear{2024}

\title{LVLM-Interpret: An Interpretability Tool for Large Vision-Language Models}

\def\authorgap{~~~}

\author{
Gabriela Ben Melech Stan\textsuperscript{1}\footnotemark[1]\authorgap
Estelle Aflalo\textsuperscript{1}\thanks{Main authors}\authorgap
Raanan Yehezkel Rohekar\textsuperscript{1}\footnotemark[1]\authorgap
Anahita Bhiwandiwalla\textsuperscript{1}\footnotemark[1]\authorgap
\\
Shao-Yen Tseng\textsuperscript{1}\footnotemark[1]\authorgap
Matthew Lyle Olson\textsuperscript{1}\footnotemark[1]\authorgap
Yaniv Gurwicz\textsuperscript{1}\footnotemark[1]\authorgap
Chenfei Wu\textsuperscript{2}\authorgap
Nan Duan\textsuperscript{2}\authorgap
Vasudev Lal\textsuperscript{1}\\
\textsuperscript{1}Intel Labs ~~~~~~
\textsuperscript{2}Microsoft Research Asia\\
\small\url{https://intellabs.github.io/multimodal_cognitive_ai/lvlm_interpret/}
}

\begin{document}
\maketitle

\begin{abstract}
In the rapidly evolving landscape of artificial intelligence, multi-modal large language models are emerging as a significant area of interest. These models, which combine various forms of data input, are becoming increasingly popular. However, understanding their internal mechanisms remains a complex task. Numerous advancements have been made in the field of explainability tools and mechanisms, yet there is still much to explore. In this work, we present a novel interactive application aimed towards understanding the internal mechanisms of large vision-language models. Our interface is designed to enhance the interpretability of the image patches, which are instrumental in generating an answer, and assess the efficacy of the language model in grounding its output in the image. With our application, a user can systematically investigate the model and uncover system limitations, paving the way for enhancements in system capabilities. Finally, we present a case study of how our application can aid in understanding failure mechanisms in a popular large multi-modal model: LLaVA.
\end{abstract}    

\section{Introduction}

Recently, large language models (LLM), such as those in the families of GPT \cite{openai2023gpt4} and LLaMA \cite{touvron2023llama, touvron2023llama2}, have demonstrated astounding understanding and reasoning capabilities, as well as the ability to generate output that adheres to human instructions.
Building on this ability, many work, such as GPT-4V \cite{openai2023gpt4v}, Qwen-VL \cite{bai2023qwen}, Gemini \cite{team2023gemini}, and LLaVA \cite{liu2024visual}, have introduced visual understanding to LLMs. 
Through the addition of a vision encoder followed by finetuning on multimodal instruction-following data, these prior work have demonstrated large vision-language models (LVLM) that are able to follow human instructions to complete both textual and visual tasks with great aptitude. 

LLMs are rapidly surpassing humans in many tasks such as summarization, translation, general question answering, and even creative writing.
However, they are still very prone to hallucination, \textit{i.e.}~the fabrication of untrue information \cite{ji2023survey,xu2024hallucination}.
This phenomenon of hallucination is also seen in LVLMs and may even include additional dimensions stemming from the visual modality \cite{wang2023evaluation, zhou2024analyzing}.
With the introduction of LVLMs and their massively increased number of parameters, interpreting and explaining model outputs to mitigate hallucination is becoming an ever rising challenge. 
In light of the need to understand the reasoning behind model responses, we present an interpretability tool for large vision-language models: LVLM-Interpret.
The proposed application adapts multiple interpretability methods to large vision-language models for interactive analysis. These methods include raw attention, relevancy maps, and causal interpretation. 
LVLM-Interpret is applicable to any LVLM with a transformer-based LLM front-end. 
We further demonstrate how we can gain insight on the inner workings of LVLMs using our application. 

The main contributions of this paper are:
\begin{itemize}
    \item We propose an interactive tool for interpreting the inner attention mechanisms of large vision-language models
    \item We present a case study that sheds light on a possible cause behind certain failure cases in LVLMs
    \item Through a study on causal explanations, we postulate that large vision-language models (such as LLaVA \cite{liu2024visual}) implicitly learn to represent causal structure 
\end{itemize}

\begin{figure}[tb]
    \centering
    \includegraphics[width=\columnwidth]{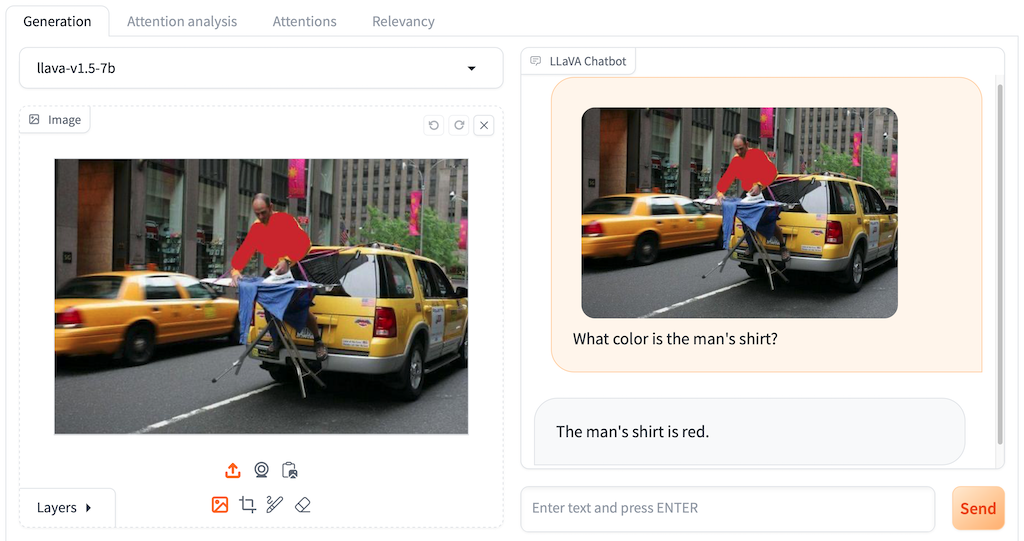}
    \caption{Main interface of LVLM-Interpret. Users can issue multimodal queries using a chatbot interface. Basic image-editing feature allows for model probing.}
    \label{fig:main_interface}
    \vspace{-1em}
\end{figure}

\begin{figure*}[t]
    \centering
    \begin{subfigure}{0.48\linewidth}
        \centering
        \includegraphics[width=\linewidth]{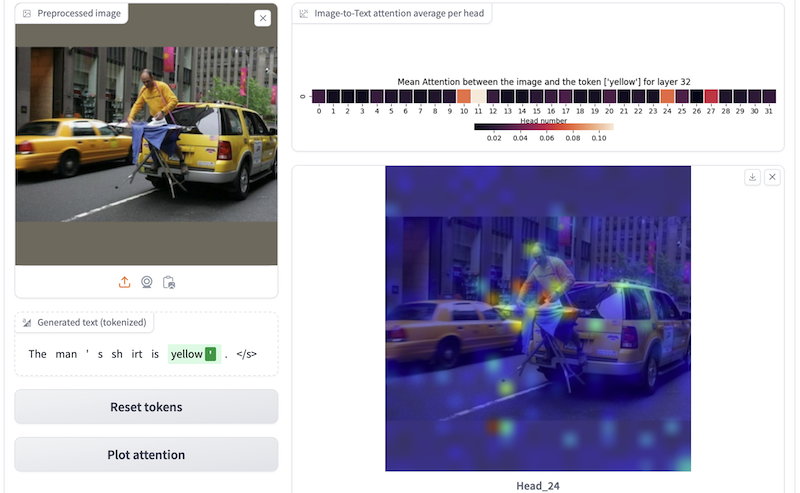}
        \caption{Image-to-Query raw attentions. The user is able to select a token or a group of tokens to visualize the attention values of image to
        text output for each head and layer.}
        \label{fig:img2text_attn}
    \end{subfigure}
    \hfill
    \begin{subfigure}{0.5\linewidth}
     \centering
        \includegraphics[width=\linewidth]{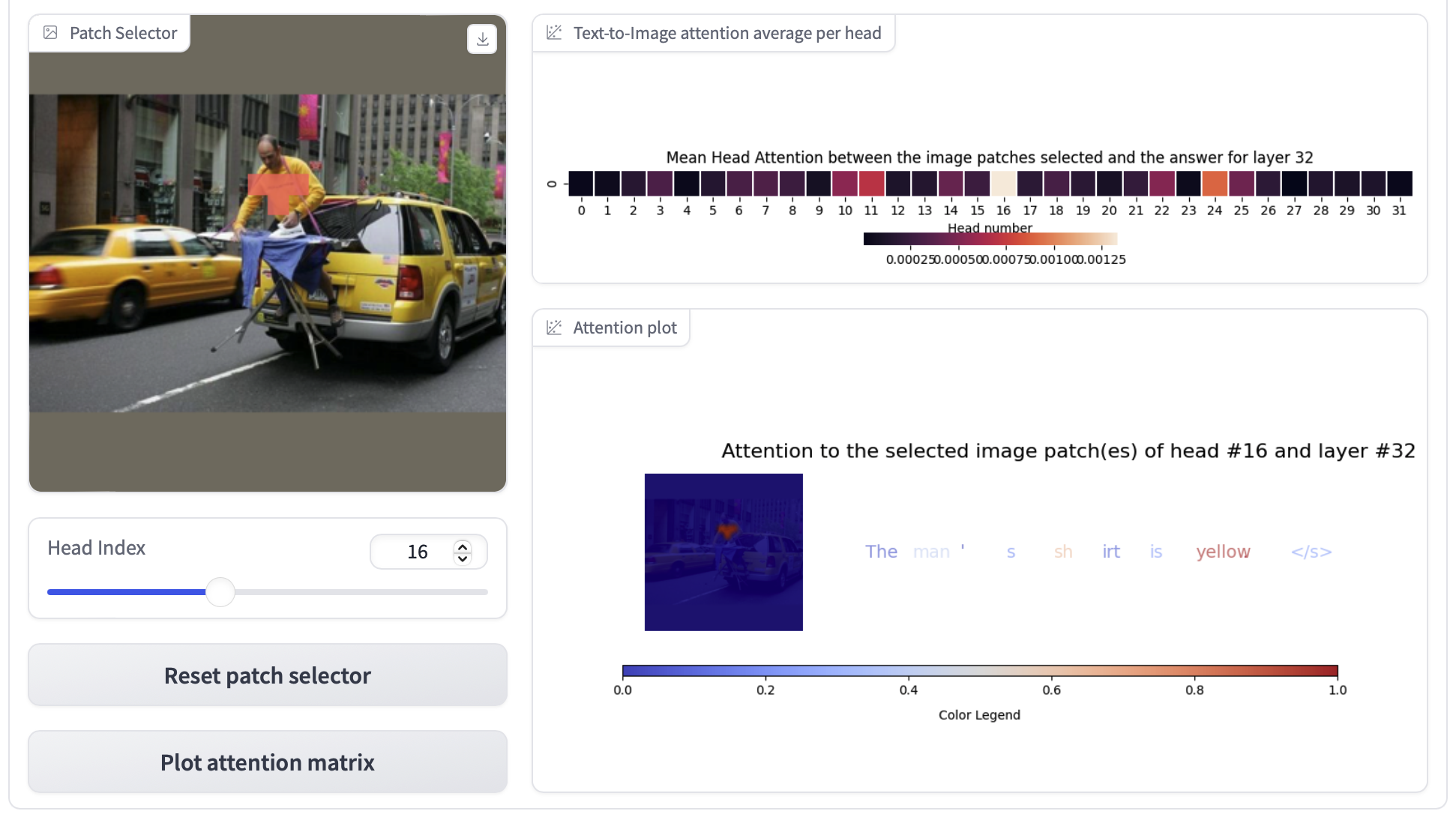}
        \caption{Query-to-Image raw attentions. The user is able to select image patches to visualize attention values going into answer tokens.}
        \label{fig:text2img_attn}
    \end{subfigure}
    \caption{Visualization of cross-modal attentions}
    \label{fig:crossmodal_attn}
    \vspace{-1em}
\end{figure*}

\section{Related Work}
The advancements in deep learning models has been preceded by novel interpretability and explainability tools to better understand the internal workings of these models. Earlier works \citep{zhang2018interpreting, zhang2019interpreting, wexler2019if} demonstrated the use of explanatory graphs, decision trees, histograms, respectively to analyze machine learning models. As Transformer \cite{vaswani2017attention} based architectures gained popularity in the field, various approaches such as \cite{chefer2021transformer} proposed computing relevancy scores across the layers of the model, \cite{rigotti2022attentionbased} generalized the attention from low-level input features to high-level concepts to ensure interpretability within a specific domain, while \cite{pan2021ia} presented a novel interpretability-aware redundancy reduction transformer framework. 

\subsection{Interpretability of Vision Models}
Studying the interpretability of Vision Transformers (ViT) has gained popularity with task-specific analysis like image captioning \cite{cornia2022explaining, sun2022explain, elguendouze2023explainability}, object detection \cite{wu2019towards, dong2022towards, baek2023swin}, image recognition \cite{mannix2024scalable, xue2022protopformer}. Recently, there has been an increased demand for interpretability analysis for the medical domain in applications such as pathology \cite{naseem2022vision, Komorowski_2023_CVPR}, retinal image classification \cite{playout2022focused, he2023interpretable} and COVID-19 analysis \cite{9642985, shome2021covid} among others. A novel vision transformer was presented in \cite{qiang2023interpretability} with a training procedure which has an interpretability-aware training objective. \cite{Nalmpantis_2023_CVPR} proposed a method to use the activations of ViT's hidden layers to predict the relevant parts of the input that contribute to its final predictions, while \cite{10132428} introduced quantification indicators to measure the impact of patch interactions to effectively exploit responsive fields of patches in ViT. 

\subsection{Interpretability of Multimodal Models}
Multimodal models have proliferated various domains across healthcare, multimedia, industrial applications among others. There has been a rise in independent interpretability studies of such multimodal systems \cite{ramesh2022investigation, aflalo2022vl, swamy2024multimodn, lyu2022dime, liu2023multimodal, Chefer_2021_ICCV}. For the medical domain where the reasoning behind decisions, high stakes involved are of utmost importance.
\cite{mallick2024ifi} demonstrated the use of an interpretability method based on attention gradients to guide the transformer training in a more optimal direction, while \cite{bi2023multimodal} presented an interpretable fusion of structural MRI and functional MRI modalities to enhance the accuracy of schizophrenia.

Our hope with this proposed interpretability tool is not to replace domain-specific solutions, but to complement them, improving existing and future large vision language models and further strengthen the confidence in the predictions and behavior of these models.

\section{Interface and Interpretability Functions }

LVLM-Interpret was developed using Gradio \cite{gradio} and follows a standard layout for multimodal chat.
Figure \ref{fig:main_interface} shows an example of the user interface.
In the UI, a user is able to upload an image and issue multimodal queries to the LVLM. 
An added editing feature allows for basic modification of the input image to probe the model with adversarial variations. 
As the LVLM model generates a response, the attention weights of the model are stored internally and are later presented to the user for visualization. 
The application also constructs relevancy maps and causal graphs relating to the output of the model.
Once a response is returned, the user is able to utilize these results as a way to interpret the model output. 
The following sections describe each of these interpretability functions.

\begin{figure*}
  \centering
  \begin{subfigure}{0.19\linewidth}
    \includegraphics[width=0.9\linewidth]{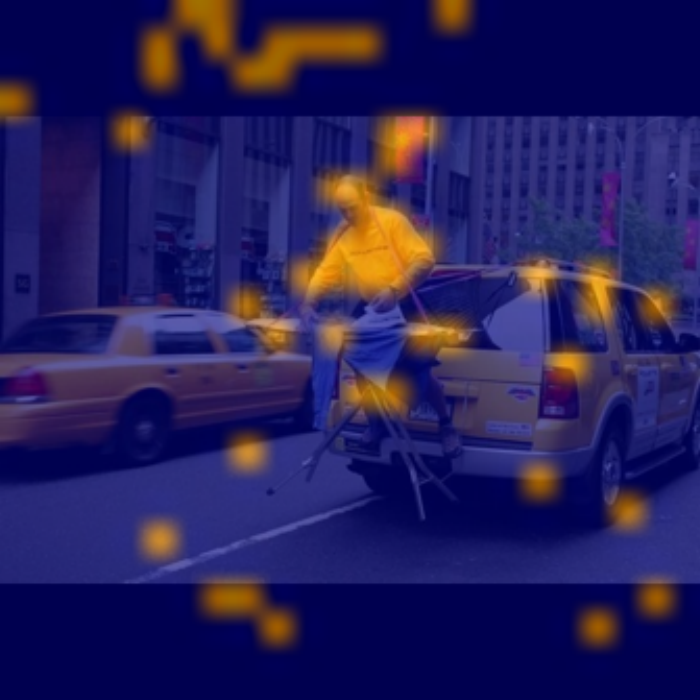}
    \caption{Raw high-Attention}
    \label{fig:raw-a}
  \end{subfigure}
  \hfill
  \begin{subfigure}{0.19\linewidth}
    \includegraphics[width=0.9\linewidth]{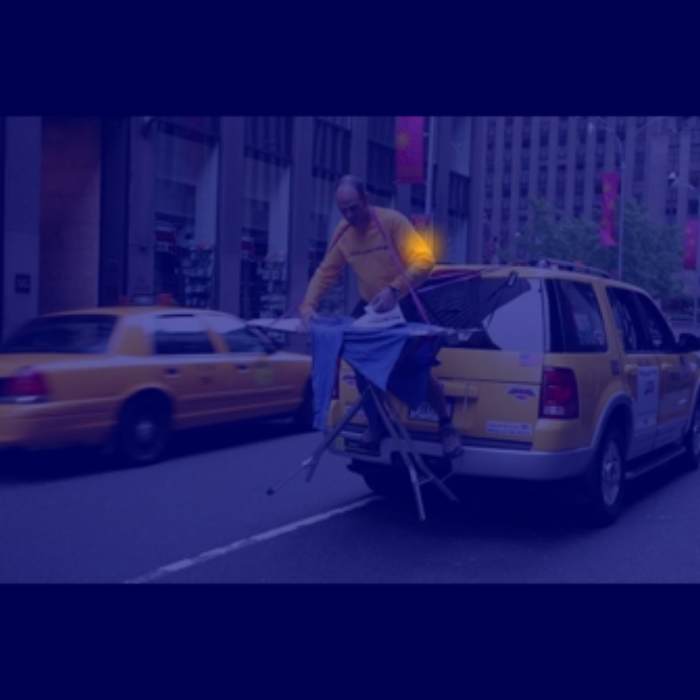}
    \caption{Search distance = 1}
    \label{fig:causal-b}
  \end{subfigure}
  \begin{subfigure}{0.19\linewidth}
    \includegraphics[width=0.9\linewidth]{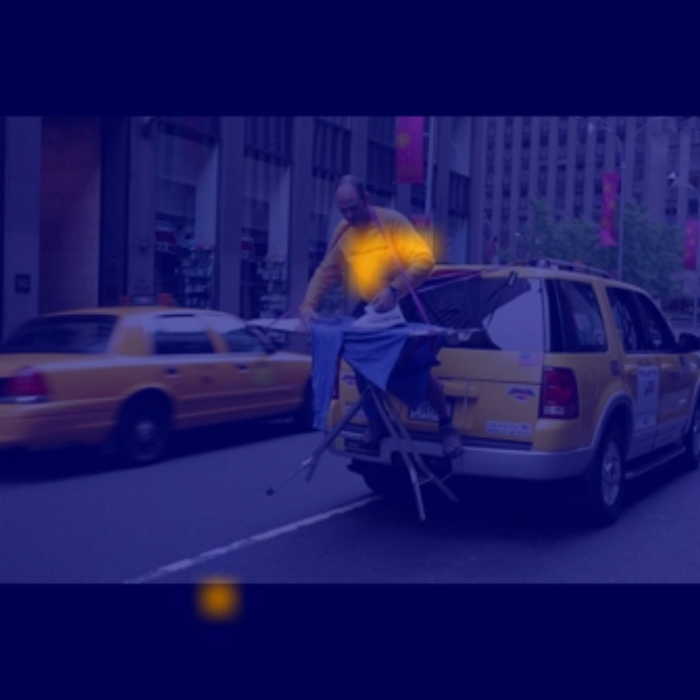}
    \caption{Search distance = 2}
    \label{fig:causal-c}
  \end{subfigure}
  \begin{subfigure}{0.19\linewidth}
    \includegraphics[width=0.9\linewidth]{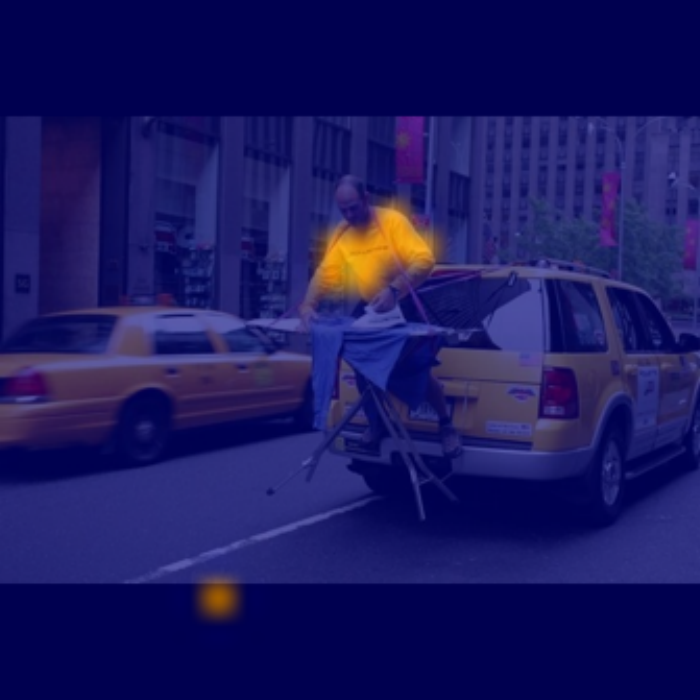}
    \caption{Search distance = 3}
    \label{fig:causal-d}
  \end{subfigure}
  \begin{subfigure}{0.19\linewidth}
    \includegraphics[width=0.9\linewidth]{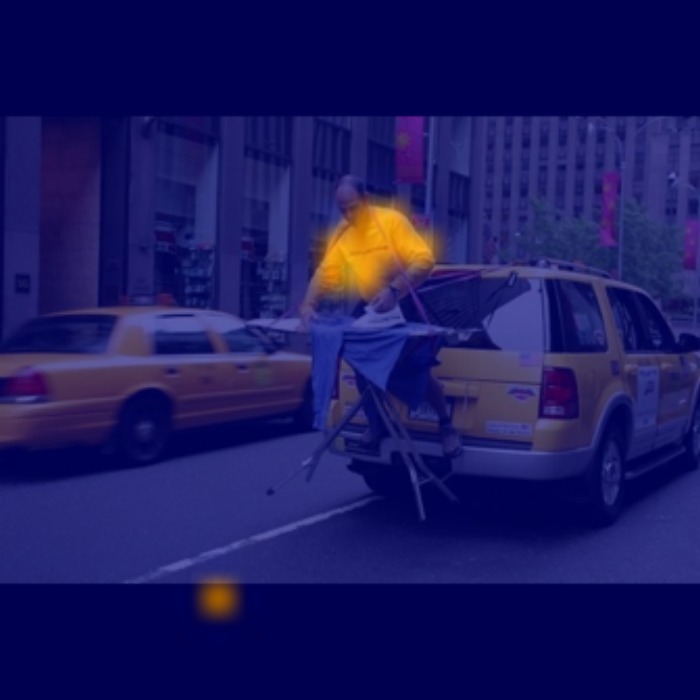}
    \caption{Search distance = 4}
    \label{fig:causal-e}
  \end{subfigure}
  \caption{Causality-based explanation for the token `\texttt{yellow}' in the generated answer `\texttt{The man's shirt is yellow}' at head 24. (a) Top 50 image-tokens having the highest raw attention values. Each serves as a graph node. (b-e) Image tokens from the explanation set identified by the CLEANN method, at different search distances on the learned causal graph. Tokens are marked with yellow blobs.}
  \label{fig:causal-explanation}
  \vspace{-1em}
\end{figure*}

\subsection{Layer Attentions}


Following work such as VL-Interpret \cite{aflalo2022vl}, LVLM-Interpret also allows for interactive visualization of raw attentions. More specifically, our application allows users to investigate the interactions among tokens from each modality. 
Heatmaps that show average attentions between image tokens and query tokens as well as answer tokens enables the user to better understand the global behavior of raw attentions. 
Figure \ref{fig:crossmodal_attn} shows how a user can visualize raw attentions for a specific head and layer. 
As shown in Figure \ref{fig:img2text_attn}, the user can select tokens from the generated response and visualize the average attentions between image patches and the selected tokens to obtain insight on how the model attends to the image when generating each token.
Conversely, Figure \ref{fig:text2img_attn} shows how a user can select image patches and visualize the degree to which each output tokens attends to that specific location. 

\subsection{Relevancy Map}


Relevancy maps \cite{chefer2021transformer, Chefer_2021_ICCV} aim at interpreting the decision-making process of transformers. These maps are designed to enhance interpretability by illustrating how different components of an input, whether text or image, are relevant to the model's generated output, overcoming some limitations of traditional attention visualization techniques.
The method assigns a local relevancy scores to each element in the input based on their contribution to the output decision. 
We refer the reader to \cite{Chefer_2021_ICCV} for more details on the approach.








We adapted the calculation of relevancy maps to LVLMs such as LLaVA. Relevancy scores are backward propagated through the LLM as well as vision transformer commonly used as the vision encoder. 
For image analysis, the relevancy scores corresponding to image patches are reshaped into a grid that matches the layout of the original image. This grid forms the basis of the relevancy map. The relevancy map is then upscaled to the original image size using bilinear interpolation, providing a visualization of the regions of the image most relevant to each generated token.

Relevancy maps can aid in model debugging, ensuring fairness, and providing explanations for inaccuracies by identifying the the most relevant parts of the input to the generated output, as demonstrated through a case study in Section \ref{sec:Case Study}.

\subsection{Causal Interpretation}

\begin{figure}
  \centering
    \includegraphics[width=0.9\linewidth]{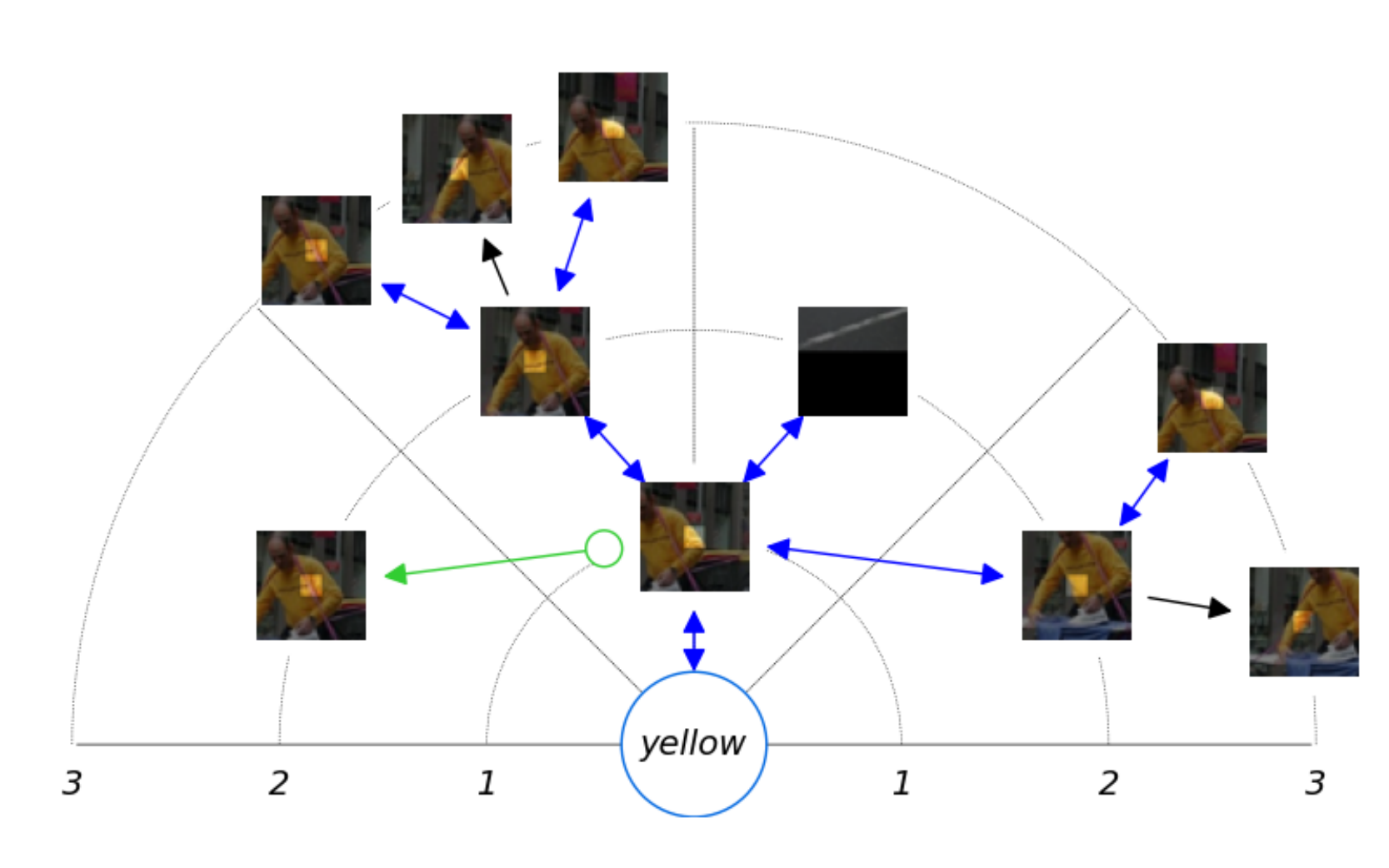}
  \caption{A tree constructed from the causal graph from which explanations for the token '\texttt{yellow}' are extracted. Arc radius indicates distance on the causal graph. Edges are color coded, bi-directed edges indicate a latent confounder, a circle edge-mark indicates that both a `tail' and 'arrow' are valid.}
  \label{fig:causal-pi-tree}
  \vspace{-1em}
\end{figure}


Recently, a causal interpretation of the attention mechanism in transformers was presented \cite{rohekar2024causal}. This interpretation leads to a method for deriving causal explanations from attention in neural networks (CLEANN). In this sense, if explanation tokens would have been masked in the input, the model would have generated a different output. Such explanations, which are a subset of the input tokens, are generally tangible and meaningful to humans. The method was previously demonstrated for a single modality, such as recommendation systems \cite{nisimov2022clear}, and text sentiment classification \cite{rohekar2024causal}. Here, we enable examining if multi-modal LLMs, which are significantly larger, internally represent causal structures. Hence, in addition to providing causal explanation, we plot the causal graphs around the explained tokens, and allow the user to decrease or increase the explanation set size based on this graph. 
We refer the reader to \cite{rohekar2021iterative, rohekar2024causal} for more details.


We employ CLEANN to explain large vision-language models by learning causal structures over input-output sequences of tokens. The presence of each token in this sequence is an event which is represented by a node in the causal graph.
Thus, the event of generating an output token is explained by the presence of a subset of input tokens. 

Consider the following example. A sequence of image token is given as part of the prompt, and the text prompt is `\texttt{What color is the man's shirt?}'. In response, the model generates `\texttt{The man's shirt is yellow.}'. One may be interested in understanding which image tokens are responsible for the the token \texttt{yellow}. That is, identify the parts of the image such that if masked, will cause the model to output a different color. 
First, the top-\textit{k} tokens having the highest attention values for \texttt{yellow} are assigned to nodes. Then, using the full attention matrix of the last (deepest) layer is used to learn a causal graph. This causal graph is a partial ancestral graph \cite{richardson2002ancestral,zhang2008completeness}, where a circle edge-mark indicates a non identifiable edge mark (head and tail are equally valid). From this graph, a tree rooted at node \texttt{yellow} is extracted (\Cref{fig:causal-pi-tree}) such that it includes all paths that potentially influence the root \cite[Appendix B. Definition 2]{rohekar2024causal}.
CLEANN searches for the minimal explaining set by gradually increasing the search distance in this tree (radius in \Cref{fig:causal-pi-tree}) from the explained node. An example is given in \Cref{fig:causal-explanation}. In \Cref{fig:raw-a}, the 50 tokens having the highest attention values are marked. In \Cref{fig:causal-b}--\Cref{fig:causal-e}, tokens within different search distances from the explained token are marked. 


This explanation approach solely relies on attention values in the last layer. While raw attention values describe pair-wise, marginal dependence relations, the causal-discovery algorithm in CLEANN identifies conditional independence relations. Thus, based only on the current trained weights, it can identify those tokens that if perturbed may change the generated token.

\begin{figure}[t]
    \centering
        \centering
        \includegraphics[width=1\linewidth]
        {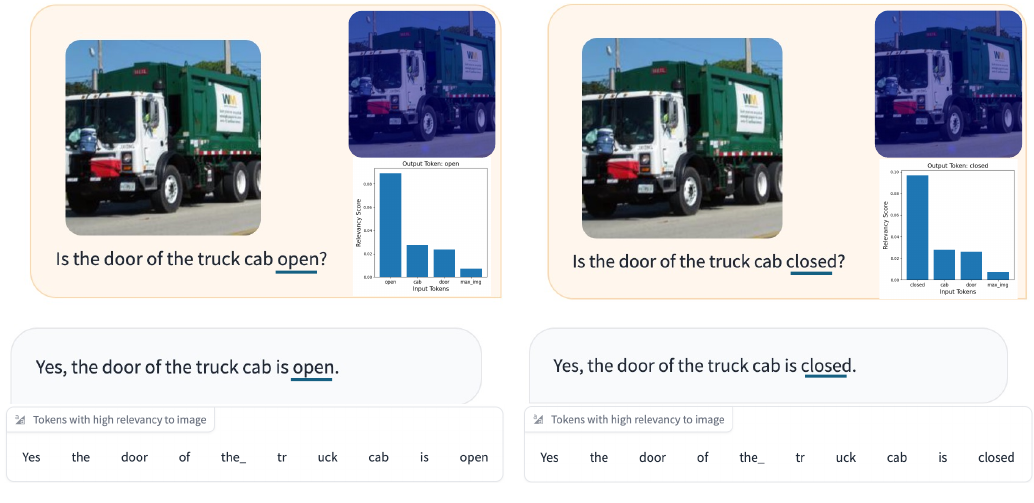}
        \caption{Example where LLaVA seems to prioritize text input over image content. Presented with an unchanging image of a garbage truck, the model provides contradictory responses (`\texttt{yes, the door is open}' vs. `\texttt{yes, the door is closed}') based on the query's phrasing. Relevancy maps and bar plots for \texttt{open} and \texttt{closed} tokens demonstrate higher text relevance compared to image.}
        \label{fig:a}
        \vspace{-1em}
    
    \vspace{0.1cm}
    
    
    \label{fig:relevancy_wrong_answers}
\end{figure}

\begin{figure}
  \centering
    \includegraphics[width=1\linewidth]{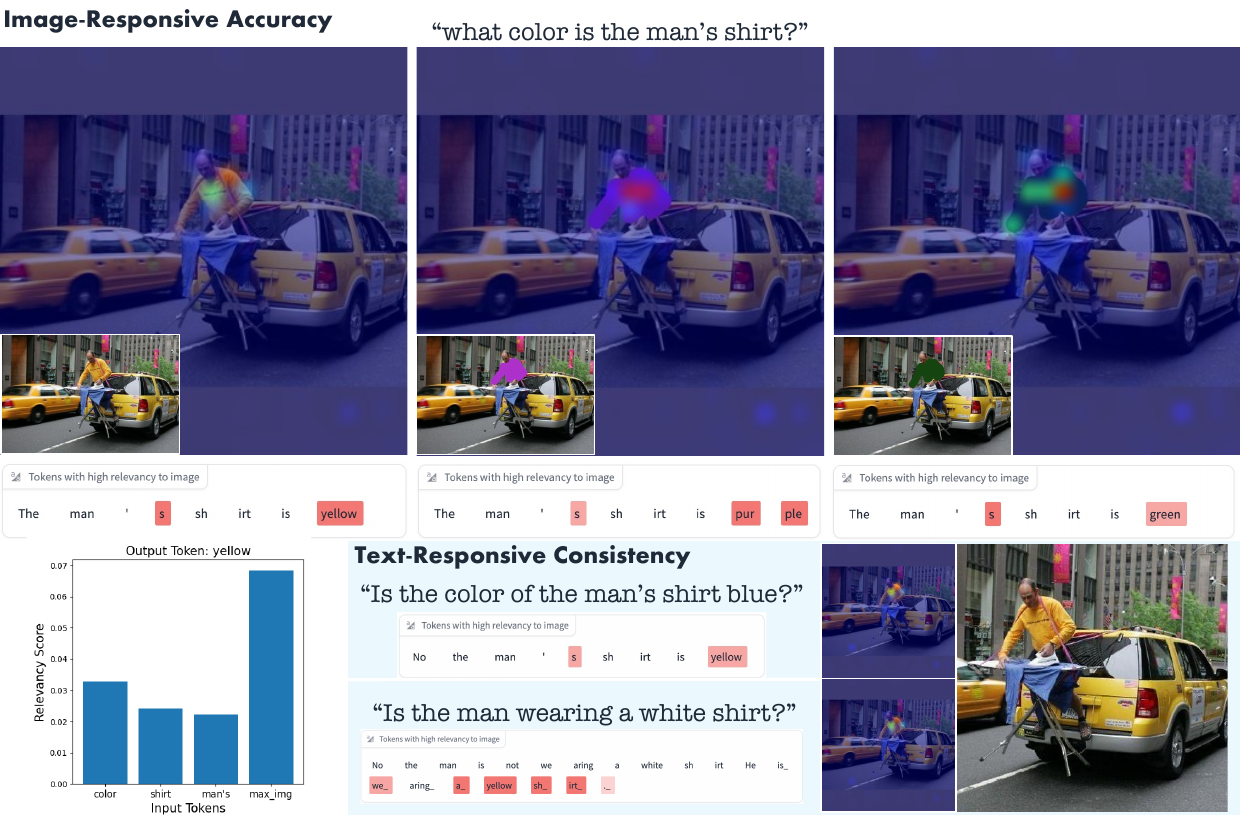}
  \caption{Example where LLaVA demonstrates visual consistency with high relevancy scores for correct output tokens. With a constant query, `\texttt{What color is the man’s shirt?}', manually altering the shirt's color in the image (purple, green) changes the model's answer which align with image changes. Relevancy scores highlight stronger connections to image than text tokens, illustrated by the image relevancy maps (upper row) and a bar plot comparison of relevancy scores. With a constant image, despite different question phrasings, the model demonstrates consistency in its answer, underscoring a strong relevancy of the \texttt{yellow} token to the visual input over the textual input variations.}
  \label{fig:relevancy_consistency_yellow_shirt}
  \vspace{-1em}
\end{figure}

\section{Case Study}
\label{sec:Case Study}


To demonstrate the functionalities of LVLM-Interpret, we analyze the LLaVA model on samples from the Multimodal Visual Patterns (MMVP) benchmark dataset \cite{tong2024eyes}. MMVP focuses on identifying ``CLIP-blind" images that are demonstrably hard for LVLMs to reason on. This dataset is of particular interest for our study since it highlights the challenges faced in answering relatively straightforward questions, often leading to incorrect responses.  

We adapted the relevancy scores to examine the impact of both text and image tokens on the output generated by LLaVA-v1.5-7b. Given that the text and vision transformers responsible for generating the input embedding were kept frozen during LLaVA finetuning, our initial step involved calculating the relevancy scores for each generated output relative to the input features to LLaMA, focusing on the LLaMA self-attention layers. We observed instances where LLaVA mainly attends to the text tokens and less to the image tokens, indicated by lower relevancy scores to the image tokens relatively to relevancy scores to the input text tokens. In these cases, the model becomes more susceptible to manipulation, in some cases altering its responses based on the query with low regard to the image content. This phenomenon is exemplified by the truck scenario depicted in Figure \ref{fig:relevancy_wrong_answers}. Conversely, when the generated outputs exhibit a greater relevance to image tokens than to input text, the accuracy of LLaVA appears to remain unaffected by how the question is phrased, as illustrated in Figure \ref{fig:relevancy_consistency_yellow_shirt}.

\section{Conclusions and Future Directions}
In this paper we presented LVLM-Interpret, an interactive tool for interpreting responses from large vision-language models. 
The tool offers a way to visualize how generated outputs relate to the input image through raw attention, relevancy maps, and causal interpretation. 
Through the many interpretability functions, users can explore the inner mechanisms of LVLMs and obtain insights on failure cases. 
The application can also reveal several potential paths for enhancing the performance of LVLMs.
Future work can include consolidation of the multiple interpretability methods for a more comprehensive metric to explain the reasoning behind model responses. 

{
    \small
    \bibliographystyle{ieeenat_fullname}
    \bibliography{main}
}


\end{document}